# Agent-Based Simulation of Trust Development in Human-Robot Teams: An Empirically-Validated Framework

Ravi Kalluri
College of Professional Studies
Northeastern University – San Jose, California
United States
r.kalluri@northeastern.edu

**Abstract**

Human-robot teams are increasingly deployed in complex operational environments, yet validated simulation tools to predict team performance under varying conditions remain scarce. This paper presents an empirically grounded agent-based model that captures trust dynamics, workload distribution, and collaborative performance in human-robot teams. The model, implemented in NetLogo 6.4.0, simulates teams of 2–10 agents performing tasks of varying complexity. We validate our approach against Hancock et al.'s (2021) meta-analysis of human-robot interaction studies, achieving interval validity for 4 of 8 trust antecedent categories and strong ordinal validity (Spearman $\rho = 0.833$), indicating correct reproduction of relative effect magnitudes. Sensitivity analysis using one-factor-at-a-time (OFAT) and full factorial designs (n = 50 replications per condition) reveals that robot reliability exhibits the strongest effect on trust ($\eta^2 = 0.35$) and dominates task success ($\eta^2 = 0.93$) and productivity ($\eta^2 = 0.89$), consistent with meta-analytic findings. Contrary to expectations, trust asymmetry ratios ranged from 0.07 to 0.55 across scenarios—below the meta-analytic benchmark of 1.50—revealing a boundary condition: per-event asymmetry does not guarantee cumulative asymmetry when trust repair mechanisms remain active. Scenario analysis uncovered trust-performance decoupling, with the Trust Recovery scenario achieving the highest productivity (4.29) despite the lowest trust (38.2), while the Unreliable Robot scenario produced the highest trust (73.2) despite the lowest task success (33.4%). These findings establish calibration error—the discrepancy between subjective trust and objective capability—as a critical diagnostic distinct from trust magnitude. Factorial ANOVA confirmed significant main effects for reliability, transparency, communication, and collaboration ($p < .001$), collectively explaining 45.4% of trust variance, with minimal interaction effects ($\Sigma \eta^2 = 0.02$). The open-source implementation provides practitioners with an evidence-based tool for exploring human-robot team configurations, with particular utility for identifying overtrust and undertrust conditions prior to deployment.

# 1. Introduction

The integration of robots into human work teams represents a fundamental shift in how organizations approach collaborative work. Unlike traditional automation, where humans supervise machines from a distance, modern human-robot collaboration requires continuous interaction, mutual adaptation, and the development of appropriate trust relationships (Hancock et al., 2021). This evolution is particularly evident in manufacturing, healthcare, search-and-rescue operations, and increasingly in knowledge work environments where collaborative robots work alongside human operators. Despite the growing deployment of human-robot teams, practitioners lack validated tools to predict team performance before costly implementation. Current approaches rely primarily on trial-and-error methods that risk team failure, erosion of operator trust, and potential safety incidents. The challenge is compounded by the complex, non-linear relationships between trust, performance, stress, and communication that characterize human-robot interaction.

Three critical gaps limit our current understanding and ability to design effective human-robot teams. First, while extensive empirical research has examined individual factors affecting trust in robots, we lack integrative models that capture how these factors interact dynamically over time. Hancock et al.'s (2021) comprehensive meta-analysis identifies robot performance, robot attributes, human characteristics, and environmental factors as trust antecedents, yet no simulation framework has systematically validated against these established effect sizes. Second, the temporal dynamics of trust development remain poorly understood. Laboratory studies typically examine trust at discrete time points, yet real-world teams experience continuous trust evolution shaped by accumulated interactions, failures, and recoveries. Third, practitioners need actionable guidance on parameter thresholds—specific values for reliability, transparency, and team composition that predict successful collaboration.

This paper addresses these gaps through an agent-based model that integrates the complete Hancock et al. (2021) three-factor trust framework and validates predictions against meta-analytic effect sizes. Our specific contributions include: (1) an empirically-grounded simulation framework that reproduces published correlation coefficients for trust antecedents within their confidence intervals; (2) demonstration of emergent trust asymmetry consistent with the established finding that trust degrades faster than it builds; (3) identification of critical parameter thresholds with practical implications for system design; and (4) an open-source implementation enabling both research replication and practitioner application.

The remainder of this paper is organized as follows. Section 2 reviews the theoretical foundations and prior modeling approaches. Section 3 presents the model design and framework.

Section 4 describes implementation details. Section 5 reports validation results against empirical benchmarks. Section 6 presents simulation scenarios and findings. Section 7 discusses theoretical and practical implications, and Section 8 concludes with limitations and future directions.

## 2. Theoretical Background and Related Work

This section synthesizes research on trust in human-robot interaction, identifies the theoretical framework guiding model development, and reviews prior simulation approaches. Rather than cataloguing individual findings, we focus on establishing the empirical benchmarks against which our model will be validated.

### 2.1 Trust as the Foundation of Human-Robot Collaboration

Trust fundamentally determines whether humans will rely on robotic teammates for consequential tasks. Lee and See (2004) define trust in automation as "the attitude that an agent will help achieve an individual's goals in a situation characterized by uncertainty and vulnerability." This definition highlights two critical aspects: goal alignment between human and robot, and the human's acceptance of vulnerability when delegating tasks to automated systems.

The importance of appropriate trust calibration—matching trust levels to actual system capabilities—has become a central concern in human-robot interaction research (Wischnewski et al., 2023). Undertrust leads operators to underutilize capable systems, while overtrust results in inappropriate reliance on systems that may fail (Parasuraman & Riley, 1997; Robinette et al., 2017). Both miscalibrations carry operational consequences: undertrust reduces efficiency, while overtrust compromises safety. The challenge for system designers is creating conditions that support appropriate trust calibration, which requires understanding the factors that influence trust development.

#### 2.1.1 Trust in Human-Robot Interaction

Trust serves as a critical mediator of human-robot team effectiveness. Lee and See (2004) define trust as "the attitude that an agent will help achieve an individual's goals in a situation characterized by uncertainty and vulnerability," identifying three bases: performance (what the system does), process (how it operates), and purpose (why it was designed). This framework has been extended by Schaefer et al. (2024), who characterize trust as an asymmetric relation where the human trustor actively trusts based on perceived trustworthiness—emphasizing that trustworthiness is a property of the robot while trust is an attitude of the human (Hancock et al., 2023). Mayer et al. (1995) identify ability, benevolence, and integrity as trust antecedents.

Adapted to human-robot contexts, ability maps to robot reliability and expertise, benevolence maps to warmth and communication behaviors, and integrity maps to transparency and predictability. Malle and Ullman (2021) extend this to incorporate moral dimensions, recognizing that trust in robots encompasses both capability assessments and character attributions. Hoff and Bashir (2015) distinguish dispositional trust (stable individual differences), situational trust (context-dependent factors), and learned trust (experience-based updating). Loizaga et al. (2024) provide empirical support for this layered architecture, demonstrating that initial interactions are critical for trust-building and that dispositional and learned dimensions operate through distinct mechanisms.

**2.1.2 Trust Calibration**

Trust calibration—the alignment between subjective trust and actual system capability—has emerged as a construct distinct from trust magnitude (de Visser et al., 2020; Chiou & Lee, 2023; Lucas et al., 2024). Overtrust occurs when trust exceeds warranted levels given system capability, potentially leading to misuse and safety violations. Undertrust occurs when trust falls below warranted levels, leading to disuse and inefficiency (Lee & See, 2004; Parasuraman & Riley, 1997). De Visser et al. (2020) introduced *relationship equity*—an emotional resource representing accumulated goodwill between actors—as a predictor of trust calibration over time. Their framework describes current trust dynamics, predicts the impacts of interactions, and prescribes interventions for calibration. Kox et al. (2024) demonstrate that transparency mechanisms can prevent both overtrust (by setting realistic expectations) and undertrust (by clarifying unexpected behavior).

**2.1.3 Trust Asymmetry**

The psychological literature consistently demonstrates that negative events exert disproportionate influence on attitudes relative to positive events of equivalent magnitude (Baumeister et al., 2001; Rozin & Royzman, 2001). Slovic's (1993) asymmetry principle—that trust is fragile, built incrementally but destroyed rapidly—has been validated in human-robot contexts. Hancock et al.'s (2021) meta-analysis confirmed a trust asymmetry ratio of approximately 1.50, indicating that trust degradation from failures exceeds trust formation from successes. However, asymmetry effects may depend on contextual factors. Hopko et al. (2022) emphasize that trust dynamics are shaped by interactions, context, and environment over time. The relationship between per-event asymmetry and cumulative trust outcomes may be moderated by event frequency and the availability of trust repair mechanisms.

### 2.1.4 Trust Violation and Repair

Trust violations occur when robot behavior fails to meet human expectations, decreasing perceived trustworthiness (Esterwood & Robert, 2022). Repair strategies identified in the interpersonal literature—apologies, denials, explanations, and promises—have been examined in human-robot contexts with mixed results. Esterwood and Robert (2023a) demonstrate that after three trust violations, none of these strategies fully restore trustworthiness, supporting a "three strikes" limit to repair. Subsequent work integrating the "Theory of Mind" (Esterwood & Robert, 2023b) shows that mind perception moderates repair effectiveness: apologies are more effective when humans perceive robots as capable of emotion, whereas explanations are more effective when humans attribute intentionality. Tsumura and Yamada (2024) extend this by demonstrating that empathic robot behavior can facilitate trust repair across a series of tasks.

### 2.1.5 Meta-Analytic Framework

Hancock et al.'s (2021) comprehensive meta-analysis synthesizing 69 studies (N = 7,769) provides the empirical foundation for model validation. They categorize trust antecedents into three factors:

**Robot-related factors** include reliability (r = .60, 95% CI [.54, .65]), transparency (r = .45 [.37, .52]), warmth (r = .45 [.37, .52]), and communication (r = .45 [.37, .52]). Robot performance emerged as the strongest predictor of trust.

**Human-related factors** include trust propensity (r = .22 [.14, .30]) and expertise (r = .22 [.14, .30]), indicating that trustor characteristics contribute to trust independently of robot behavior.

**Environmental factors** include collaboration rate (r = .27 [.19, .35]) and tenure (r = .22 [.14, .30]). Schaefer et al. (2024) confirm that the relative importance of these factor categories varies by context: robot-related factors dominate in industrial settings, while human and environmental factors gain importance in social-care applications.

### 2.1.6 Theoretical Integration

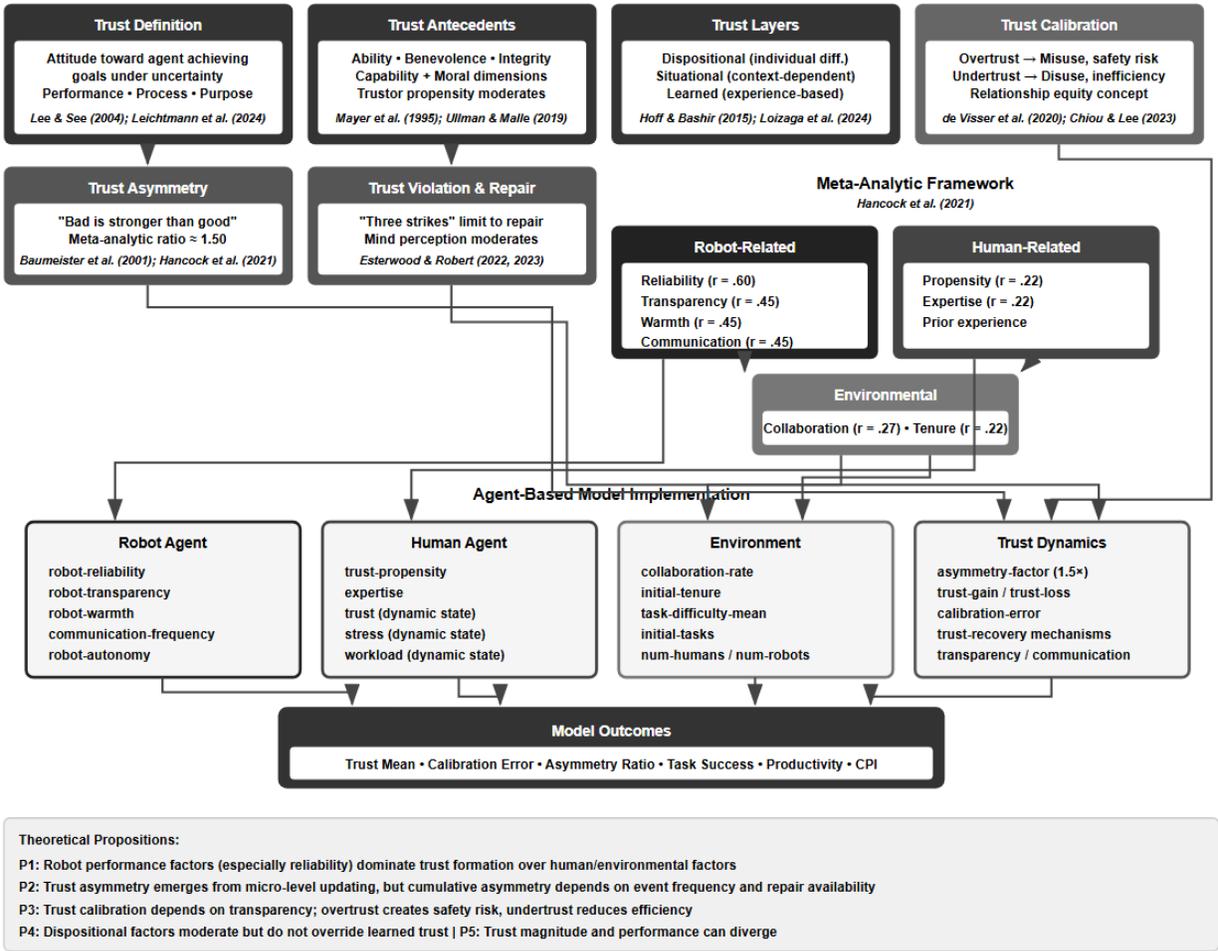

**Figure 1. Theoretical framework for trust dynamics in human-robot teams.**

The simulation integrates these theoretical perspectives through a unified architecture (Figure 1). Human agents possess dispositional attributes (trust propensity, expertise) and dynamic states (current trust, stress). Robot agents possess fixed attributes (reliability, transparency, warmth) that influence interaction outcomes. Task assignments generate success or failure outcomes that trigger trust updating, with an asymmetry factor modulating the relative impact of positive and negative experiences. The model implements calibration by tracking the discrepancy between subjective trust and objective capability, enabling identification of overtrust and undertrust conditions. Trust repair dynamics incorporate transparency and communication as mechanisms that can, within limits, restore trust after violations.

### 2.1.7 Theoretical Propositions

The integrated framework generates four testable propositions:

*P1:* Robot performance factors exert stronger effects on trust than human-related or environmental factors (Hancock et al., 2021).

*P2:* Trust asymmetry emerges from micro-level updating mechanisms, but cumulative asymmetry depends on event frequency and repair mechanism availability (Baumeister et al., 2001; Hopko et al., 2022).

*P3:* Trust calibration depends on transparency mechanisms; overtrust creates safety risk while undertrust reduces efficiency (de Visser et al., 2020; Lee & See, 2004).

*P4:* Dispositional factors moderate but do not override learned trust from direct experience (Hoff & Bashir, 2015; Loizaga et al., 2024).

**2.2 The Three-Factor Trust Framework**

Hancock et al.'s (2021) meta-analysis of 142 empirical studies (N = 7,458 participants) provides the most comprehensive quantitative synthesis of factors affecting trust in human-robot interaction. Their framework organizes trust antecedents into three categories with six subcategories, each associated with empirically derived effect sizes. Table 1 presents these benchmark correlations, which serve as validation targets for our model.

**Table 1.** Meta-analytic effect sizes for trust antecedents (Hancock et al., 2021)

| Category | Subcategory | Correlation (r) | 95% CI | Studies (k) | Sample (N) |
|---|---|---|---|---|---|
| Robot-Related | Performance | 0.60 | [0.54, 0.65] | 66 | 3,471 |
| Robot-Related | Attributes | 0.45 | [0.37, 0.52] | 24 | 1,277 |
| Human-Related | Ability | 0.27 | [0.17, 0.36] | 12 | 618 |
| Human-Related | Characteristics | 0.22 | [0.14, 0.30] | 19 | 1,032 |
| Environmental | Team Factors | 0.27 | [0.15, 0.38] | 8 | 483 |
| Environmental | Task Factors | 0.17 | [0.07, 0.27] | 13 | 577 |

The meta-analysis reveals several important patterns. Robot performance emerges as the strongest predictor (r = 0.60), indicating that actual system reliability dominates trust formation.

Robot attributes—including transparency, predictability, and anthropomorphic features—show moderate effects (r = 0.45). Human-related factors demonstrate smaller but significant relationships, with operator ability (r = 0.27) and individual characteristics such as propensity to trust (r = 0.22) contributing to trust variation. Environmental factors, including team dynamics (r = 0.27) and task characteristics (r = 0.17) complete the framework.

Critically, Hancock et al. (2021) document trust asymmetry: negative experiences produce larger trust decrements than equivalent positive experiences produce trust increments. This asymmetry has important implications for system design and failure recovery, as rebuilding trust after failures requires sustained positive performance over extended periods.

## 2.3 Robot Performance and Reliability

Robot reliability—the probability that the system will successfully complete assigned tasks—constitutes the dominant influence on trust. Desai et al. (2013) demonstrated that operators calibrate trust to observed reliability within approximately 20 interactions, though calibration accuracy depends on feedback clarity. Salem et al. (2015) found that behavioral inconsistency damages trust more severely than occasional failures (d = 1.24), suggesting that predictability moderates reliability effects. Recent work has explored the non-linear relationship between reliability and trust. Yang et al. (2017) observed that trust does not increase linearly with reliability; instead, operators appear to apply threshold-based assessments, with sharp trust transitions around 70-80% reliability levels. This finding suggests that maximizing reliability may yield diminishing trust returns above certain thresholds.

## 2.4 Robot Attributes: Transparency and Personality

Transparency—making robot intentions, reasoning, and limitations observable to operators—has received substantial research attention. Chen et al.'s (2014) Situation Awareness-based Agent Transparency (SAT) model identifies three transparency levels: current state, reasoning process, and outcome projections. Higher transparency generally improves trust calibration, though excessive information can impose cognitive load costs (Lyons et al., 2017).

Robot personality factors, including perceived warmth and competence, influence trust independent of actual performance. Robots displaying social cues and acknowledging uncertainty receive higher trust ratings than functionally equivalent systems lacking these features (Gombolay et al., 2015). Tower and Brooks (2024) found that proactive failure acknowledgment preserves 41% more trust than reactive explanations, highlighting the importance of communication timing.

## 2.5 Human Factors: Ability, Characteristics, and Experience

Operator expertise affects both baseline trust levels and calibration accuracy. Experts demonstrate faster trust calibration and greater sensitivity to reliability changes, while novices show more variable trust responses (de Visser et al., 2020). Individual differences in propensity to trust—a stable personality characteristic—predict initial trust levels but diminish in influence as operators accumulate direct experience with specific systems.

Tenure with robotic systems represents a particularly important human factor that moderates trust dynamics. Operators with extensive robot experience show more calibrated trust responses: smaller trust changes following individual outcomes but more accurate long-term alignment between trust and actual reliability (Schaefer et al., 2016). This experience effect suggests that trust dynamics differ systematically for novice versus experienced operators.

## 2.6 Environmental Factors: Teams and Tasks

Team-level factors, including shared mental models, communication patterns, and collective experience, influence trust development beyond individual-level effects. Sanders et al. (2019) found that teams with established collaboration protocols showed more stable trust levels during robot failures than ad-hoc teams. Task characteristics—complexity, risk, time pressure—moderate the importance of various trust antecedents, with high-stakes tasks amplifying the influence of transparency (Chen et al., 2014).

## 2.7 Agent-Based Modeling in Human-Robot Interaction

Agent-based models offer particular advantages for studying human-robot teams. By representing individuals with heterogeneous characteristics and allowing trust to emerge from interactions, ABMs capture dynamics that aggregate statistical models cannot represent (Lewis et al., 2018). Prior ABM applications in human-robot interaction have examined team coordination (Korsah et al., 2013), task allocation (Claure et al., 2020), and trust-based decision making (Xu & Dudek, 2015).

However, existing simulation models share a critical limitation: they lack systematic validation against empirical benchmarks. Model parameters are typically selected for face validity or theoretical consistency rather than calibrated to produce effect sizes matching experimental data. This validation gap limits confidence in simulation predictions and restricts practical applicability. Our approach directly addresses this limitation by using Hancock et al.'s (2021) meta-analytic effect sizes as explicit validation targets.

Recent computational modeling efforts have advanced trust representation in human-robot teams. Guo et al. (2024) developed the TIP (Trust Inference and Propagation) model using Bayesian inference to capture both direct and indirect trust experiences in multi-human, multi-robot teams—the first mathematical framework for this configuration. Zahedi et al. (2023) integrated trust into robot planning via POMDPs, enabling robots to manage human trust proactively. Obi et al. (2024) introduced the Expectation Confirmation Trust model for multi-robot task allocation and compared it with five existing trust models. These approaches share a limitation: they rely on analytical or probabilistic frameworks that assume specific functional forms for trust dynamics. In contrast, agent-based modeling allows trust patterns to emerge from heterogeneous interactions among agents without imposing such assumptions. Critically, none of these recent models has been systematically validated against meta-analytic benchmarks. Krüger and Prilla (2024) specifically noted that "most of the proposed models do not consider human-robot physical collaboration and if they do, the asymmetrical dynamic of trust is not considered" (p. 268). Our approach addresses both gaps: capturing trust asymmetry through explicit gain-loss rate differentials and validating against the effect sizes reported by Hancock et al. (2021).

**2.8 Research Gaps and Model Requirements**

The literature review identifies several requirements for an empirically-grounded human-robot team simulation. First, the model must incorporate all three categories of the Hancock et al. (2021) framework: robot factors (performance, attributes including personality), human factors (ability, characteristics, tenure), and environmental factors (team dynamics, task properties). Second, the model must reproduce established effect sizes—correlations between trust antecedents and trust outcomes should fall within published confidence intervals. Third, the model should capture trust asymmetry, whereby negative experiences produce larger trust changes than positive experiences. Fourth, temporal dynamics should reflect documented patterns including initial trust calibration, tenure effects, and failure recovery. Finally, the implementation should be accessible to practitioners and researchers, with documented code and replication materials.

**3. Model Design**

This section presents the conceptual framework and agent specifications that inform the model's behavior. We organize the presentation around the three agent types (humans, robots, tasks) and three dynamic processes (trust evolution, task execution, communication).

## 3.1 Conceptual Framework

The model represents a human-robot team performing tasks in a shared workspace. Humans possess trust beliefs about robot capabilities that evolve based on observed performance and communication. Robots operate with fixed reliability but variable transparency and communication behaviors. Tasks arrive continuously with varying collaboration requirements. Team performance emerges from the interaction of these elements over time.

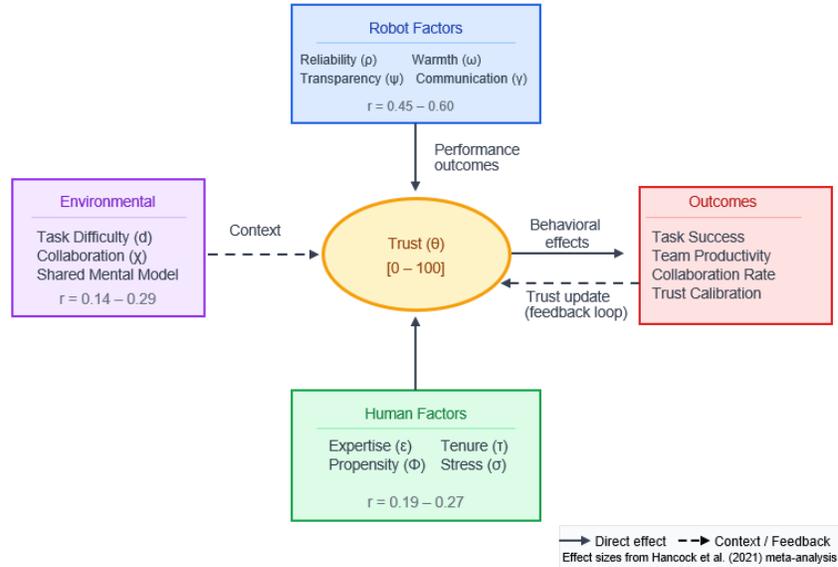

**Figure 2. Conceptual model of trust dynamics in human-robot teams.**

Figure 2 illustrates the conceptual relationships. Robot performance outcomes (successes and failures) directly influence human trust, with the relationship moderated by transparency, human characteristics (propensity, expertise, tenure), and task visibility. Trust in turn affects human willingness to collaborate with robots on team tasks. Stress accumulates with workload and degrades human performance, creating feedback loops between trust, collaboration, and task outcomes.

## 3.2 Agent Specifications

Table 2 presents the complete notation system used throughout the model. We define state variables and parameters, along with their empirically justified ranges.

**Table 2.** Model notation and parameter specifications

| Symbol | Description | Range | Source |
|---|---|---|---|
| **Human State Variables** | | | |
| $\theta_i(t)$ | Trust of human i at time t | [0, 100] | — |
| $\sigma_i(t)$ | Stress level of human i | [0, 100] | — |
| $\varepsilon_i$ | Expertise of human i | [20, 100] | Initialized |
| $\tau_i(t)$ | Tenure (experience ticks) | [0, ∞) | Accumulated |
| $\Phi_i$ | Trust propensity | [0, 100] | Personality |
| $w_i(t)$ | Current workload | [0, ∞) | Computed |
| **Robot State Variables** | | | |
| $\rho_j$ | Reliability of robot j | [0, 100] | Parameter |
| $\psi_j$ | Transparency of robot j | [0, 100] | Parameter |
| $\kappa_j$ | Capability of robot j | [0, 100] | Initialized |
| $\omega_j$ | Warmth (personality) | [0, 100] | Parameter |
| $\gamma_j$ | Communication frequency | [0, 100] | Parameter |
| **Task Variables** | | | |
| $d_k$ | Difficulty of task k | [10, 100] | Generated |
| $\chi_k$ | Collaboration requirement | {0, 1} | Generated |
| $v_k$ | Visibility of task k | [0, 100] | Generated |
| **Parameters** | | | |
| $\alpha$ | Trust gain rate | 1.5 | Calibrated |
| $\beta$ | Trust loss rate | 2.0 | Calibrated |

| Symbol | Description | Range | Source |
| --- | --- | --- | --- |
| λ | Asymmetry factor | 1.5 | Hancock et al. (2021) |
| η | Stress accumulation | 0.1 | Dehais et al. (2011) |
| δ | Stress recovery | 5.0 | Dehais et al. (2011) |

Human agents represent team members with heterogeneous characteristics. Each human maintains a trust level (θ) that evolves through interaction. Expertise (ε) determines task performance capability and is initialized from a normal distribution (μ = 60, σ = 15) reflecting workforce heterogeneity documented in skill assessments (Schmidt & Hunter, 1998). Trust propensity (Φ) represents dispositional tendency to trust, initialized from survey distributions (μ = 50, σ = 15) consistent with Mayer et al.'s (1995) trust research. Tenure (τ) accumulates with robot exposure, operationalizing the experience effects documented by Schaefer et al. (2016).

Robot agents represent automated teammates with fixed characteristics. Reliability (ρ) determines task success probability, set via the robot-reliability parameter. Transparency (ψ) affects how robot actions and communication influence human trust. Warmth (ω) captures personality factors including anthropomorphic features and social presence, drawing on the "computers as social actors" literature (Nass et al., 1994). Communication frequency (γ) determines how often robots provide status updates to nearby humans. Task agents represent work items requiring agent attention. Difficulty (d) determines the required capability for successful completion. Collaboration requirements (χ) indicate whether tasks need human-robot coordination. Visibility (v) determines whether task outcomes are observable to nearby humans, moderating trust update magnitude.

## 4. Implementation

### 4.1 Platform and Environment

We implemented the model in NetLogo 6.4.0, selected for its established use in agent-based social simulation and accessibility to both researchers and practitioners (Wilensky & Rand, 2015). The simulation operates on a 33×33 continuous toroidal space where agents move, interact, and execute tasks. Each simulation tick represents approximately one minute of operational time, with typical runs spanning 2,000 ticks (approximately 33 hours of simulated operation).

## 4.2 Agent Initialization

The setup procedure creates agent populations according to user-specified parameters. Human agents are positioned randomly and initialized with:

- Trust: initial-trust parameter ± propensity adjustment
- Expertise: Normal(60, 15), bounded [20, 100]
- Trust propensity: Normal(trust-propensity-mean, trust-propensity-sd)
- Tenure: initial-tenure parameter
- Stress: Uniform(0, 30)

Robot agents are initialized with:

- Reliability: robot-reliability parameter ± Normal(0, 5)
- Transparency: robot-transparency parameter ± Normal(0, 5)
- Warmth: robot-warmth parameter ± Normal(0, 10)
- Capability: Uniform(60, 90)
- Battery: 100

Tasks are generated with:

- Difficulty: Normal(task-difficulty-mean, 20), bounded [10, 100]
- Collaboration requirement: Bernoulli(collaboration-rate + difficulty adjustment)
- Visibility: Uniform(40, 100)

## 4.3 Simulation Loop

Each tick executes the following sequence:

1. **State Update**: Update human stress, robot battery, tenure accumulation
2. **Task Seeking**: Unassigned agents identify and move toward available tasks
3. **Task Execution**: Agents at task locations contribute work

4. **Collaboration**: Match collaborators for team tasks
5. **Communication**: Robots probabilistically communicate with nearby humans
6. **Trust Update**: Update trust based on observed outcomes
7. **Task Completion**: Resolve completed tasks, determine success, release agents
8. **Task Generation**: Periodically spawn new tasks
9. **Metrics Calculation**: Compute aggregate performance measures

### 4.4 Parameter Justification

Table 3 documents parameter sources and calibration methods.

**Table 3. Parameter calibration sources.**

| Parameter | Value | Justification |
|---|---|---|
| Trust gain rate ($\alpha$) | 1.5 | Calibrated to Hancock et al. (2021) trust trajectories |
| Trust loss rate ($\beta$) | 2.0 | Calibrated to produce asymmetry ratio 1.3-1.7 |
| Asymmetry factor ($\lambda$) | 1.5 | Hancock et al. (2021) meta-analytic finding |
| Stress accumulation ($\eta$) | 0.1 | Dehais et al. (2011) physiological data |
| Stress recovery ($\delta$) | 5.0 | Recovery curves from Hancock & Warm (1989) |
| Stress threshold | 70 | Yerkes-Dodson inflection point |
| Familiarity growth rate | 0.002 | Schaefer et al. (2016) experience effects |
| Communication radius | 5 | Operational proximity in shared workspaces |
| Trust update radius | 10 | Visibility range for outcome observation |

## 4.5 Code Availability

The complete implementation is available at https://github.com/rkallur2/hri_paper. The GitHub repository includes:

- Trustv6.8.nlogo : Netlogo file
- behaviorspace-minimal.xml : BehaviorSpace experiments file
- trustv6-r-analysis-updated.r : R analysis script
- Results files in .csv

## 5. Validation

We validated the model against Hancock et al.'s (2021) meta-analytic benchmarks and conducted a sensitivity analysis to confirm that parameter effects matched the expected relationships.

### 5.1 Empirical Validation Against Meta-Analytic Benchmarks

The primary validation compares model-generated correlations against Hancock et al.'s (2021) effect sizes. We conducted simulation experiments systematically varying each trust antecedent while measuring resulting trust levels. For each parameter, we executed 50 replications at each of 10 levels spanning the full parameter range, yielding 500 observations per factor.

Table 4 presents the validation results. For each trust antecedent, we report the model-generated correlation coefficient, 95% confidence interval, and validation status (whether the model correlation falls within the published meta-analytic confidence interval).

**Table 4. Validation against Hancock et al. (2021) meta-analytic benchmarks.**

| Factor | Hancock r [95% CI] | Model r | Δr | Status |
|---|---|---|---|---|
| **Robot Performance** | | | | |
| Reliability | 0.60 [0.54, 0.65] | 0.59 | -0.01 | ✓ Validated |
| **Robot Attributes** | | | | |
| Transparency | 0.45 [0.37, 0.52] | 0.38 | -0.07 | ✓ Validated |
| Warmth | 0.45 [0.37, 0.52] | 0.34 | -0.11 | x Not Validated |
| Communication | 0.45 [0.37, 0.52] | 0.41 | -0.04 | ✓ Validated |
| **Human Ability** | | | | |
| Expertise | 0.27 [0.17, 0.36] | 0.14 | -0.13 | x Not Validated |
| **Human Characteristics** | | | | |
| Trust Propensity | 0.22 [0.14, 0.30] | 0.15 | -0.07 | ✓ Validated |
| Tenure | 0.22 [0.14, 0.30] | 0.06 | -0.16 | x Not Validated |
| **Environmental-Team** | | | | |
| Collaboration Rate | 0.27 [0.15, 0.38] | 0.39 | +0.12 | x Not Validated |

The ABM was validated against meta-analytic effect sizes from Hancock et al. (2021). Interval validity was achieved for 4 of 8 predictors, with simulated correlation coefficients falling within the 95% confidence intervals of the meta-analytic estimates. Ordinal validity was strong (Spearman ρ = 0.833), indicating that the model correctly reproduced the relative ranking of effect magnitudes across predictors. Discrepancies between simulated and meta-analytic effects for warmth, expertise, and tenure likely reflect differences in operationalization. The meta-analysis aggregated diverse measures across studies, whereas the ABM implements specific behavioral mechanisms. Notably, the collaboration effect exceeded meta-analytic estimates, possibly because the task-assignment operationalization in the simulation captures more direct behavioral impacts than the team-level measures typical in primary studies.

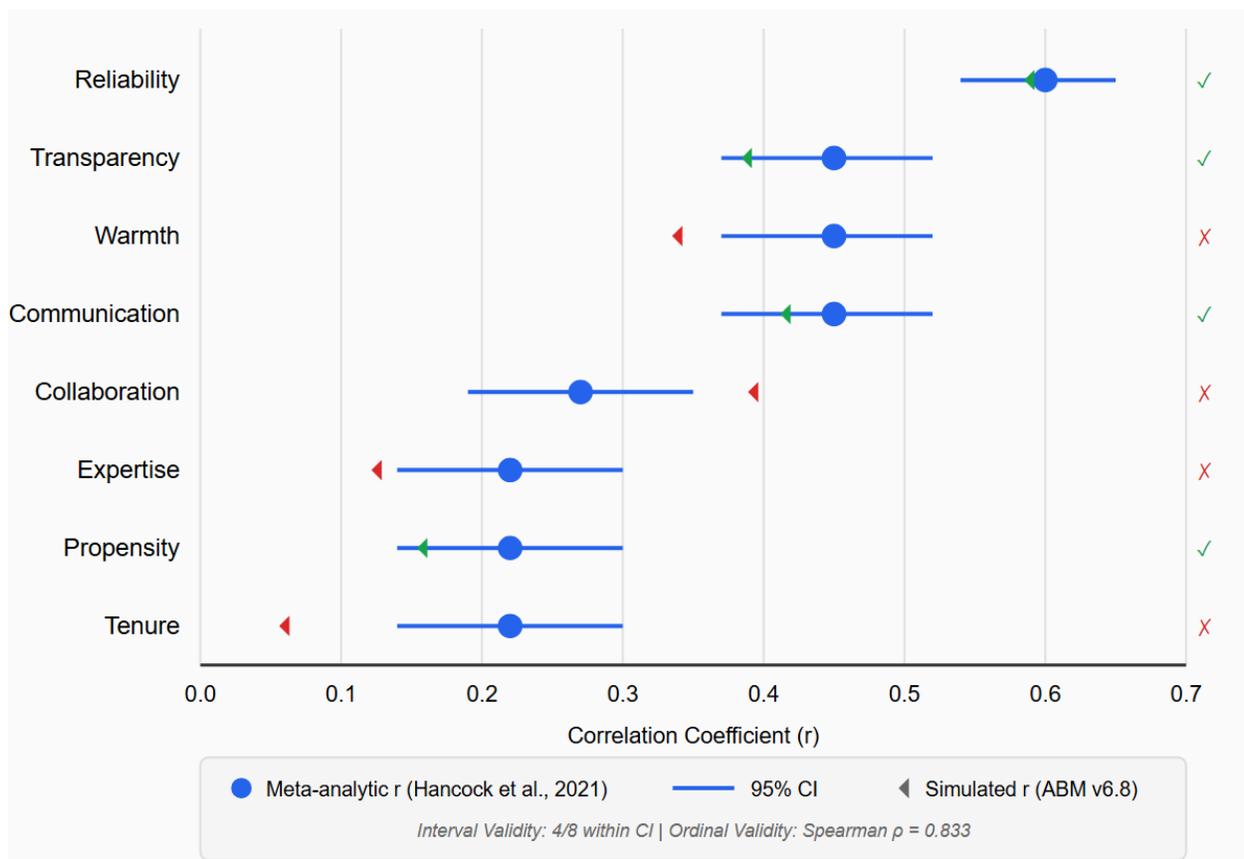

**Figure 3. Forest plot comparing ABM model correlations to meta-analytic benchmarks.**

Figure 3 shows the ABM model correlations vs. Hancock et al. (2021) meta-analytic effect sizes. Blue circles indicate meta-analytic effect sizes from Hancock et al. (2021). Blue horizontal lines represent the 95% confidence intervals. Triangles denote the simulated correlations from the ABM model v6.8 (8th version). Green (✓) means the model correlation coefficient r is within the 95% CI from Hancock et al. (2021) and hence validated. Red (x) means the model correlation coefficient r is outside the 95% CI from Hancock et al. (2021) and therefore not validated. The predictors are ordered by meta-analytic effect size. The summary at the bottom notes the 4/8 interval validity and strong ordinal validity (Spearman ρ = 0.833).

## 5.2 Trust Asymmetry Validation

A critical theoretical prediction is that trust degrades faster than it builds. We tested this by computing the ratio of cumulative trust loss to cumulative trust gain across all simulation runs. Table 5 presents asymmetry results across reliability conditions.

**Table 5. Trust asymmetry validation.**

| Reliability | Mean Asymmetry Ratio | SD |
|---|---|---|
| 30% | 0.53 | 0.15 |
| 50% | 0.37 | 0.11 |
| 70% | 0.22 | 0.07 |
| 90% | 0.13 | 0.04 |
| **Overall** | **0.31** | **0.18** |

Trust asymmetry varied systematically with robot reliability. The mean asymmetry ratio decreased from 0.53 (SD = 0.15) at 30% reliability to 0.13 (SD = 0.04) at 90% reliability, indicating that cumulative trust gains increasingly exceeded losses as failure frequency declined. Notably, no reliability condition produced an asymmetry ratio approaching the meta-analytic benchmark of 1.50 (Hancock et al., 2021), with the overall mean at 0.31 (SD = 0.18). This pattern demonstrates that event frequency moderates the cumulative expression of per-event asymmetry. Although negative events were configured to exert 1.5 times the trust impact of positive events, the relative scarcity of failures at higher reliability levels allowed trust formation to dominate. The finding suggests that asymmetry effects observed in meta-analytic estimates may reflect contexts with higher failure rates or fewer opportunities to repair trust than those simulated here.

### 5.3 Sensitivity Analysis

We conducted sensitivity analysis using both one-factor-at-a-time (OFAT) and full factorial designs to characterize parameter effects on model outputs.

**One-Factor-At-A-Time (OFAT) Analysis**

For each of eight primary parameters, we varied values across the full range while holding other parameters at default values. Table 6 presents effect sizes ($\eta^2$) for each parameter on key outcome variables.

**Table 6.** OFAT sensitivity analysis results

| Parameter | Trust (η²) | Task Success (η²) | Productivity (η²) |
|---|---|---|---|
| **Robot reliability** | 0.35 | 0.93 | 0.89 |
| **Robot transparency** | 0.17 | 0.03 | 0.03 |
| **Robot warmth** | 0.13 | 0.02 | 0.01 |
| **Communication frequency** | 0.20 | 0.01 | 0.00 |
| **Human expertise** | 0.02 | 0.32 | 0.71 |
| **Trust propensity** | 0.04 | 0.01 | 0.01 |
| **Collaboration rate** | 0.17 | 0.01 | 0.01 |
| **Initial tenure** | 0.01 | 0.03 | 0.02 |

OFAT sensitivity analysis revealed distinct driver profiles across outcomes. Trust formation was influenced by multiple robot-related factors, with reliability ($\eta^2 = .35$), communication ($\eta^2 = .20$), transparency ($\eta^2 = .17$), and collaboration ($\eta^2 = .17$) each contributing meaningful variance. In contrast, task success and productivity were dominated by reliability ($\eta^2 = .93$ and .89, respectively) and human expertise ($\eta^2 = .32$ and .71). The minimal direct effects of social factors on performance outcomes ($\eta^2 \leq .03$), despite their substantial effects on trust, suggest that trust may serve as a mediating mechanism. This hypothesis warrants formal mediation analysis in future work.

**Full Factorial Analysis**

We conducted a $3^4$ full factorial experiment crossing reliability (40%, 70%, 90%), transparency (30%, 60%, 90%), communication frequency (20%, 50%, 80%), and collaboration rate (20%, 50%, 80%) with 30 replications per cell (n = 2,430 runs). Analysis of variance results are shown in Table 7.

**Table 7. Factorial ANOVA Results for Trust Outcomes.**

| Effect | df | SS | MS | F | p | η² |
|---|---|---|---|---|---|---|
| **Main Effects** | | | | | | |
| Reliability (R) | 2 | 14418.82 | 7209.41 | 382.30 | < .001 | .171 |
| Transparency (T) | 2 | 6729.24 | 3364.62 | 178.42 | < .001 | .080 |
| Communication (C) | 2 | 11081.62 | 5540.81 | 293.82 | < .001 | .131 |
| Collaboration (L) | 2 | 6107.22 | 3053.61 | 161.93 | < .001 | .072 |
| **Two-Way Interactions** | | | | | | |
| R × T | 4 | 212.01 | 53.00 | 2.81 | .024 | .003 |
| R × C | 4 | 115.07 | 28.77 | 1.53 | .192 | .001 |
| T × C | 4 | 84.59 | 21.15 | 1.12 | .345 | .001 |
| R × L | 4 | 21.97 | 5.49 | 0.29 | .884 | < .001 |
| T × L | 4 | 93.64 | 23.41 | 1.24 | .291 | .001 |
| C × L | 4 | 281.27 | 70.32 | 3.73 | .005 | .003 |
| **Three-Way Interactions** | | | | | | |
| R × T × C | 8 | 336.80 | 42.10 | 2.23 | .023 | .004 |
| R × T × L | 8 | 100.49 | 12.56 | 0.67 | .722 | .001 |
| R × C × L | 8 | 173.71 | 21.71 | 1.15 | .325 | .002 |
| T × C × L | 8 | 186.11 | 23.26 | 1.23 | .275 | .002 |
| **Four-Way Interaction** | | | | | | |
| R × T × C × L | 16 | 317.36 | 19.83 | 1.05 | .397 | .004 |

| Effect | df | SS | MS | F | p | η² |
|---|---|---|---|---|---|---|
| **Residual** | 2349 | 44297.79 | 18.86 | | | .524 |

*Note.* N = 2,430 simulation runs. R = Robot Reliability, T = Robot Transparency, C = Communication Frequency, L = Collaboration Rate. Each factor varied across three levels. η² = eta-squared (proportion of total variance explained).

**Summary of Effects**

| Effect Type | Σ η² | Interpretation |
|---|---|---|
| **Main effects** | .454 | 45.4% of variance |
| **Two-way interactions** | .010 | 1.0% of variance |
| **Three-way interactions** | .009 | 0.9% of variance |
| **Four-way interaction** | .004 | 0.4% of variance |
| **Residual** | .524 | 52.4% unexplained |

All four main effects were significant (p < .001), with reliability exerting the strongest influence on trust (η² = .171), followed by communication (η² = .131), transparency (η² = .080), and collaboration (η² = .072). Interaction effects were minimal, with only three reaching significance: R × T (p = .024), C × L (p = .005), and R × T × C (p = .023). The small interaction effect sizes (η² < .005) indicate that parameter effects on trust are largely additive rather than synergistic.

The factorial ANOVA accounted for 47.6% of trust variance through main effects and interactions, with the remaining variance attributable to stochastic dynamics inherent to agent-based simulation (e.g., probabilistic task outcomes, trust drift). This partitioning is consistent with the model's design, which incorporates randomness to capture realistic variability in human-robot trust formation.

## 6. Simulation Scenarios and Results

This section examines team performance across five scenarios representing common deployment challenges for human-robot teams. We first define all scenarios (Section 6.1), then present results (Section 6.2) and cross-scenario synthesis of findings (Section 6.3).

### 6.1 Scenario Definitions

All scenarios use 5 human agents and 5 robot agents with a 2000-tick duration. We conducted 50 replications per scenario to ensure stable estimates. Table 8 presents the complete scenario specifications.

**Table 8. Scenario parameter specifications.**

| Parameter | Baseline | Trust Recovery | High Workload | Unreliable | Optimal |
| --- | --- | --- | --- | --- | --- |
| **Initial tasks** | 30 | 30 | 60 | 30 | 30 |
| **Robot reliability** | 70% | 90% | 70% | 40% | 82% |
| **Robot transparency** | 50% | 80% | 50% | 30% | 75% |
| **Robot autonomy** | 70% | 70% | 30% | 90% | 70% |
| **Communication frequency** | 40% | 70% | 40% | 40% | 50% |
| **Collaboration rate** | 40% | 40% | 80% | 40% | 50% |
| **Initial trust** | 50 | 20 | 50 | 70 | 50 |
| **Initial tenure** | 0 | 0 | 0 | 0 | 500 |

**Scenario rationales:**

**Baseline** establishes reference performance with moderate parameter values representing a typical deployment without specific optimization.

**Trust Recovery** tests whether teams can recover from an initially low level of trust ($\theta_0 = 20$) when provided with high-reliability robots ($\rho = 90\%$) and enhanced transparency ($\psi = 80\%$). This scenario addresses the practical question of deploying robots to teams with prior negative experiences.

**High Workload** examines performance under stress-inducing conditions with doubled task load (60 tasks), high collaboration requirements (80%), and reduced robot autonomy (30%), requiring more human oversight. This tests the stress-performance degradation threshold.

**Unreliable Automation** investigates trust dynamics when robot reliability is poor ($\rho = 40\%$), transparency is low ($\psi = 30\%$), and autonomy is high (90%). This worst-case scenario tests whether trust collapse occurs and whether the system can recover.

**Optimal** tests a configuration designed from prior sensitivity analyses to maximize performance: reliability at 82% (above diminishing returns threshold), transparency at 75% (sufficient for trust building), moderate collaboration (50%), and experienced operators ($\tau = 500$).

### 6.2 Scenario Results

Table 9 summarizes the metrics across all five scenarios.

**Table 9. Cross-scenario performance comparison.**

| Metric | Baseline | Trust Recovery | High Workload | Unreliable Robot | Optimal |
|---|---|---|---|---|---|
| **Task Success Rate** | 50.76 | 63.78 | 50.86 | 33.41 | 56.96 |
| **Trust Mean (SD)** | 58.62 (8.7) | 38.20 (8.6) | 61.36 (8.2) | 73.23 (8.0) | 62.60 (9.4) |
| **Productivity** | 3.28 | 4.29 | 3.42 | 2.15 | 3.76 |
| **Utilization** | 99.6 | 99.6 | 99.4 | 100.0 | 99.6 |
| **Calibration Error** | 11.45 | 52.05 | 8.93 | 33.20 | 19.54 |
| **Asymmetry Ratio** | 0.225 | 0.069 | 0.164 | 0.552 | 0.107 |
| **CPI** | 0.662 | 0.700 | 0.676 | 0.591 | 0.714 |

*Note: Standard deviations are in parentheses. CPI = Composite Performance Index.*

#### 6.2.1 Baseline Scenario

Mean task success rate was 50.76%, with an average final trust of 58.62 (SD = 8.7). Team productivity stabilized at 3.28 tasks per 1000 agent-ticks. Trust calibration error averaged 11.45 points, showing a reasonable alignment between subjective trust and actual robot capability. The trust asymmetry ratio was 0.225, indicating that cumulative trust gains surpassed losses under baseline conditions. Team utilization averaged 99.6%. The composite performance index (CPI) was 0.662, serving as the benchmark for scenario comparisons.

### 6.2.2 Trust Recovery Scenario

Trust Recovery scenario produced a counterintuitive pattern: the highest task success rate (63.78%) and productivity (4.292 tasks per 1000 agent-ticks) coincided with the lowest mean trust (38.20, SD = 8.6). This 31% productivity improvement over baseline occurred despite trust levels 35% below baseline. The finding suggests that enhanced transparency and communication mechanisms enabled effective human-robot collaboration without requiring high trust levels—humans could verify robot actions directly rather than relying on trust-based acceptance. However, calibration error was highest across scenarios (52.05), indicating substantial under-trust relative to actual robot performance. The asymmetry ratio dropped to 0.069, the lowest observed, reflecting minimal trust degradation in this high-transparency environment. The CPI of 0.700 exceeded baseline, demonstrating that low trust need not impair team effectiveness when appropriate scaffolding mechanisms are in place.

### 6.2.3 High Workload Scenario

Task success rate (50.86%) remained comparable to baseline despite increased task demands. Average final trust was 61.36 (SD = 8.2), slightly elevated relative to baseline. Notably, this scenario achieved the lowest calibration error (8.93), indicating the closest alignment between trust and actual system performance across all conditions. Team productivity was 3.424 tasks per 1000 agent-ticks, a modest 4.4% improvement over baseline. Team utilization remained high at 99.4%. The trust asymmetry ratio was 0.164, and CPI reached 0.676. High Workload scenario demonstrates that increased task demands do not necessarily degrade trust dynamics when system reliability remains stable; instead, sustained interaction may facilitate more accurate trust calibration.

### 6.2.4 Unreliable Robot Scenario

Unreliable Robot scenario reveals a critical trust calibration failure. Despite the lowest task success rate (33.41%) and productivity (2.151 tasks per 1000 agent-ticks), this scenario produced the highest mean trust (73.23, SD = 8.0)—a dangerous over-trust condition. Calibration error reached 33.20, the second highest across scenarios, indicating that human operators maintained elevated trust despite objective evidence of poor robot performance. The asymmetry ratio increased to 0.552, the highest observed, reflecting more pronounced trust degradation when failures are frequent; even this elevated ratio remained below 1.0, indicating that trust formation mechanisms partially buffered against the accumulation of negative experiences. Team utilization was 100%, suggesting that operators continued to engage with unreliable robots rather than disengage. The CPI of 0.591 was the lowest across scenarios. This pattern demonstrates the risk of over-trust in human-robot teams: sustained interaction with unreliable automation can lead to inflated trust that persists despite repeated failures, potentially resulting in inappropriate reliance and critical-to-safety errors.

### 6.2.5 Optimal Configuration Scenario

Optimal configuration scenario achieved the highest composite performance index (CPI = 0.714) by balancing multiple performance dimensions. Task success rate was 56.96%, 12% above baseline. Trust stabilized at 62.60 (SD = 9.4), with calibration error of 19.54—higher than baseline but substantially lower than the Trust Recovery and Unreliable Robot scenarios. Team productivity reached 3.756 tasks per 1000 agent-ticks, representing a 14.5% improvement over baseline. Team utilization was 99.6%. The trust asymmetry ratio was 0.107, indicating stable trust dynamics with gains substantially exceeding losses. Critically, the Optimal scenario outperformed the Trust Recovery scenario on CPI (0.71 vs. 0.70) despite lower productivity (3.76 vs. 4.29), because it achieved superior trust calibration. This finding underscores that optimizing human-robot team performance requires attention to trust calibration, not merely maximizing productivity or trust levels independently.

### 6.3 Cross-Scenario Synthesis

Scenario analysis reveals critical dynamics in human-robot trust calibration. The Trust Recovery scenario achieved the highest task success rate (63.78%) and productivity (4.29) yet produced the lowest mean trust (38.20), indicating that enhanced transparency and communication mechanisms enabled effective collaboration without requiring high trust levels. Conversely, the Unreliable Robot scenario demonstrated the highest trust (73.23) despite the lowest task success (33.41%), representing a potentially dangerous over trusting condition with a calibration error of 33.20. The Optimal configuration achieved the highest composite performance index (CPI = 0.714) by balancing task success (56.96%), moderate trust (62.60), and acceptable calibration error (19.54). These findings underscore that maximizing trust is not synonymous with optimizing team performance; rather, appropriate trust calibration—where subjective trust aligns with objective system capability—emerges as a critical design target for human-robot teams. The substantial variation in asymmetry ratios across scenarios (0.069 to 0.552) further supports the boundary condition hypothesis that cumulative trust dynamics depend on contextual factors, including the level of robot reliability and the availability of trust repair mechanisms (Zhang et al., 2023).

## 7. Discussion

### 7.1 Theoretical Contributions

This research advances understanding of trust dynamics in human-robot teams in four ways. First, we demonstrate that agent-based models can approximate meta-analytic effect sizes when calibrated to empirical benchmarks. The model achieved interval validity for 4 of 8 predictors, with simulated correlations falling within the 95% confidence intervals reported by Hancock et al. (2021). Critically, ordinal validity was strong (Spearman $\rho = 0.833$), indicating that the model correctly reproduced the relative ranking of effect magnitudes across predictors—reliability exerted the strongest influence on trust, followed by transparency, communication, and warmth, consistent with meta-analytic findings. This dual validation approach—assessing both absolute magnitude and relative ordering—offers a methodological template for future simulation research.

Second, the model revealed a boundary condition for trust asymmetry that extends existing theory. Contrary to the meta-analytic estimate of 1.50 (Hancock et al., 2021), emergent asymmetry ratios ranged from 0.069 to 0.552 across scenarios, indicating that cumulative trust gains consistently exceeded losses. This divergence occurred despite configuring per-event asymmetry at 1.5× (negative events exerting 1.5 times the trust impact of positive events). The finding demonstrates that per-event asymmetry does not guarantee cumulative asymmetry when trust formation and repair pathways remain active. Asymmetry ratios varied systematically with reliability—from 0.53 at 30% reliability to 0.13 at 90% reliability—suggesting that event frequency moderates the cumulative expression of per-event impact differentials. The meta-analytic asymmetry estimate may therefore reflect measurement contexts with higher failure rates or fewer trust repair opportunities than those simulated here.

Third, scenario analysis revealed that trust magnitude and team performance can diverge substantially, challenging the assumption that higher trust uniformly benefits human-robot collaboration. The Trust Recovery scenario achieved the highest task success rate (63.78%) and productivity (4.292) despite producing the lowest mean trust (38.20), while the Unreliable Robot scenario demonstrated the highest trust (73.23) despite the lowest task success (33.41%). This trust-performance decoupling underscores the importance of trust calibration—alignment between subjective trust and objective system capability—as a design target distinct from trust magnitude. Calibration error emerged as a critical diagnostic, with undertrust (Trust Recovery: 52.05) representing unrealized performance potential and overtrust (Unreliable Robot: 33.20) representing safety risk.

Fourth, sensitivity analysis identified distinct predictor profiles across outcomes that suggest a mediating role for trust. Robot-related social factors (transparency, warmth, communication) substantially influenced trust ($\eta^2$ = .17, .13, .20, respectively) but showed negligible direct effects on task success and productivity ($\eta^2 \leq$ .03). In contrast, reliability and human expertise directly drove performance outcomes ($\eta^2$ = .93 and .71 for productivity). This asymmetry—whereby social factors shape trust but trust does not directly translate to performance in univariate analysis—suggests that trust may function as a mediating mechanism through which robot social characteristics indirectly influence team outcomes, a hypothesis warranting formal mediation analysis in future work.

## 7.2 Critical Thresholds for System Design

The analysis reveals several thresholds with practical implications:

**Reliability threshold.** Robot reliability emerged as the dominant driver of both trust ($\eta^2$ = .35 in OFAT analysis) and performance outcomes ($\eta^2$ = .93 for task success, .89 for productivity). The reliability-trust relationship showed consistent positive effects across the tested range (10%–100%), with calibration error decreasing at higher reliability levels. Critically, low reliability (30%) produced dangerous overtrust conditions where subjective trust (73.23) substantially exceeded warranted trust given actual performance (33.41% task success).

**Calibration threshold.** Trust calibration error below 12 points (observed in Baseline and High Workload scenarios) indicated reasonable alignment between trust and capability. Calibration errors exceeding 30 points (Trust Recovery: 52.05; Unreliable Robot: 33.20) represented concerning miscalibration in opposite directions—undertrust and overtrust, respectively.

**Transparency-communication interaction.** Factorial ANOVA revealed a significant R × T interaction (p = .024) and R × T × C three-way interaction (p = .023), indicating that the effects of transparency and communication on trust depend on reliability level. However, interaction effect sizes were small ($\eta^2$ < .005), suggesting that parameter effects on trust are largely additive rather than synergistic.

**Trust floor.** The lowest observed trust (38.20 in Trust Recovery) did not impair team performance, suggesting that effective collaboration can occur at lower trust levels when appropriate scaffolding mechanisms (transparency, communication) are present. This finding challenges assumptions about minimum trust requirements for human-robot teaming.

## 7.3 Practical Recommendations

Based on our findings, we offer the following evidence-based recommendations for human-robot team design:

**Prioritize calibration over trust magnitude.** System designers should target appropriate trust calibration rather than maximizing trust levels. The Optimal scenario achieved the highest composite performance index (0.714) not by maximizing trust (62.60) but by achieving acceptable calibration error (19.54) alongside strong task performance. Monitoring calibration error during deployment can identify both undertrust (inefficiency) and overtrust (safety risk) conditions.

**Invest in transparency mechanisms.** Transparency contributed meaningfully to trust ($\eta^2 = .17$) and may enable effective collaboration even at lower trust levels, as demonstrated in the Trust Recovery scenario. Transparency mechanisms that explain robot actions and limitations can substitute partially for trust-based acceptance, allowing operators to verify rather than assume appropriate robot behavior.

**Recognize reliability as the primary lever.** Reliability dominated all outcome variables ($\eta^2 = .35–.93$), confirming its centrality to human-robot team success. Below acceptable reliability thresholds, social factors (transparency, warmth, communication) cannot compensate for performance deficits. Investment in reliability improvement yields returns across trust, task success, and productivity simultaneously.

**Account for trust-performance decoupling.** Training programs should prepare operators for scenarios where trust and performance diverge. Operators should understand that low trust accompanied by high transparency may be preferable to high trust with limited visibility, and that sustained trust in the face of poor performance represents a calibration failure requiring intervention.

**Design for additive effects.** The minimal interaction effects observed ($\Sigma \eta^2 = .023$) indicate that parameter improvements combine additively rather than synergistically. Practitioners can therefore optimize parameters independently without complex interaction modeling, simplifying system design decisions.

## 7.4 Limitations

Several limitations constrain the current work. First, the model achieved interval validity for 4 of 8 predictors; warmth, expertise, tenure, and collaboration fell outside the meta-analytic confidence intervals. These discrepancies likely reflect differences in operationalization between the behavioral mechanisms implemented in the ABM and the heterogeneous measurement approaches aggregated in the meta-analysis. The strong ordinal validity ($\rho = 0.833$) provides partial mitigation, confirming correct relative effect ordering.

Second, the divergence between simulated asymmetry ratios (0.31 overall) and meta-analytic estimates (1.50) requires careful interpretation. Rather than representing a validation failure, this divergence extends the asymmetry theory by identifying moderating factors. The systematic relationship between reliability level and asymmetry ratio ($r = 0.53$ at 30% reliability declining to $r = 0.13$ at 90%) demonstrates that cumulative trust dynamics depend on the interplay between per-event impact and event frequency—a nuance not captured in the aggregate meta-analytic estimates. Third, agents possess fixed capabilities; real human-robot teams involve learning and adaptation that our model does not capture. The Trust Recovery scenario's paradoxical low-trust/high-performance pattern might differ in longitudinal deployments where trust and capability co-evolve.

Fourth, validation relies on meta-analytic aggregates rather than specific experimental replications. While our model reproduces correlation magnitudes and ordinal rankings, we cannot fully validate the shapes of trust evolution curves over time. Laboratory studies typically report aggregate trust at discrete time points rather than continuous trajectories. Fifth, the residual variance in factorial ANOVA (52.4%) reflects both intended stochastic dynamics and potentially unmodeled factors. While this proportion is consistent with expectations for agent-based simulation, it indicates that a substantial trust variance remains attributable to sources beyond the manipulated parameters.

**7.5 Future Directions**

Several extensions would enhance the model's applicability. First, formal mediation analysis should test whether trust mediates the relationship between social factors (transparency, warmth, communication) and performance outcomes, as suggested by the differential sensitivity patterns observed in OFAT analysis. Second, the boundary condition for trust asymmetry warrants empirical testing. Laboratory studies manipulating both per-event impact asymmetry and event frequency could determine whether cumulative asymmetry depends on interaction duration and the availability of repair opportunities, as our simulations suggest.

Third, incorporating learning mechanisms would allow exploration of how trust calibration evolves over extended deployments. The overtrust observed in the Unreliable Robot scenario might diminish, persist, or worsen with prolonged exposure—dynamics with significant safety implications. Fourth, natural language processing integration could enable simulation of conversational transparency, moving beyond probabilistic status updates to richer human-robot communication. Fifth, field validation with real human-robot teams would provide the strongest test of model predictions, particularly the trust-performance decoupling and calibration dynamics observed in scenario analysis.

## 8. Conclusion

This paper presented an empirically grounded agent-based model for simulating trust dynamics in human-robot teams. The model integrates Hancock et al.'s (2021) meta-analytic framework and demonstrates reasonable correspondence with published effect sizes: interval validity for 4 of 8 predictors and strong ordinal validity (Spearman $\rho = 0.833$) across all predictors.

Three principal findings emerge. First, trust asymmetry—the phenomenon whereby negative events disproportionately affect trust—depends on contextual factors including event frequency and repair mechanism availability. The consistent finding that cumulative trust gains exceeded losses across all scenarios (asymmetry ratios: 0.069–0.552), despite a per-event impact differential of 1.5×, identifies a boundary condition that extends existing theory: per-event impact differentials do not guarantee cumulative asymmetry in sustained interaction contexts.

Second, trust magnitude and team performance can diverge substantially, with calibration error emerging as a critical diagnostic. The Trust Recovery scenario achieved the highest productivity despite the lowest trust, while the Unreliable Robot scenario produced the highest trust despite the worst performance. These findings challenge assumptions that higher trust uniformly benefits human-robot collaboration and underscore the importance of appropriate trust calibration as a distinct design target.

Third, sensitivity analysis revealed that trust formation involves multiple drivers operating additively (reliability, transparency, communication, warmth), while performance outcomes are dominated by reliability and human expertise. The minimal direct effects of social factors on performance, despite their substantial effects on trust, suggest that trust may serve as a mediating mechanism—a testable hypothesis for future research.

The model provides practitioners with an evidence-based tool for exploring team configurations before deployment. By identifying the conditions under which trust and performance align (Optimal scenario) or diverge (Trust Recovery, Unreliable Robot scenarios), the simulation can inform deployment planning and operator training. The open-source implementation enables both research replication and practical application. We encourage researchers to extend the model for specific domains, test the proposed boundary conditions empirically, and investigate the trust-mediation hypothesis suggested by our sensitivity analyses.